\documentclass[letterpaper, 10 pt, conference]{ieeeconf} 
\IEEEoverridecommandlockouts                              %
\usepackage{times}
\usepackage{multicol}
\usepackage[bookmarks=true]{hyperref}

\usepackage{amsmath,amsfonts,amssymb}
\usepackage[dvipsnames]{xcolor}
\usepackage{pstricks,pstricks-add}
\usepackage[noend]{algorithm2e}
\usepackage{placeins}
\usepackage{float}
\usepackage{caption}
\SetAlCapSkip{1em}
\usepackage{graphicx}
\usepackage{xcolor}

\DeclareMathOperator*{\argmax}{arg\,max}

\newenvironment{flushitemize}{%
\begin{list}{$\bullet$}
   {\setlength{\leftmargin}{15pt}}%
    \setlength{\labelwidth}{20pt}
    \setlength{\itemindent}{0pt}
    \setlength{\labelsep}{0.5em}
 \setlength{\itemsep}{1pt}
 \setlength{\parskip}{0pt} 
 \setlength{\parsep}{0pt}}
 {\end{list}}

\author{
Pietro Pierpaoli$^1$, Harish Ravichandar$^2$, Nicholas Waytowich$^3$, Anqi Li$^4$,\\ Derrik Asher$^3$, and Magnus Egerstedt$^1$
\thanks{*This work was supported by the Army Research Lab under Grant DCIST CRA W911NF-17-2-0181.}
\thanks{$^1$ School of Electrical and Computer Engineering, Georgia Institute of Technology, Atlanta, GA 30332, USA. {\tt\small \{pietro.pierpaoli,magnus\}@gatech.edu}}
\thanks{$^2$ School of Interactive Computing, Georgia Institute of Technology, Atlanta, GA 30332, USA. {\tt \small \{harish.ravichandar\}@gatech.edu}} 
\thanks{$^3$ {US Army Research Laboratory, Aberdeen Proving Grounds, MD, 21005 USA {\tt \small \{nicholas.r.waytowich, derrik.e.asher\}.civ@mail.mil}}}
\thanks{$^4$ Department of Computer Science, University of Washington, Seattle, WA 98195, USA. {\tt \small anqil4@cs.washington.edu}}
}

\begin{document}

\title{\LARGE \bf
Inferring and Learning Multi-Robot Policies from Observations}

\maketitle
\thispagestyle{empty}
\pagestyle{empty}

\begin{abstract}
We present a technique for learning how to solve a multi-robot mission that requires interaction with an external environment by observing an expert system executing the same mission. We define the expert system as a team of robots equipped with a library of controllers, each designed to solve a specific task, supervised by an expert policy that appropriately selects controllers based on the states of robots and environment. The objective is for an un-trained team of robots (i.e., imitator system) equipped with the same library of controllers, but agnostic to the expert policy, to execute the mission, with performances comparable to those of the expert system. From un-annotated observations of the expert system, a multi-hypothesis filtering technique is used to estimate individual controllers executed by the expert policy. Then, the history of estimated controllers and environmental states is used to train a neural network policy for the imitator system. Considering a perimeter protection scenario on a team of differential-drive robots, we show that the learned policy endows the imitator system with performances comparable to those of the expert system.
\end{abstract}

\section{Introduction} \label{sec:intro}


Motivated by the benefits of robots capable to collaborate (e.g., resilience to faults~\cite{ramachandran2019resilience,pierpaoli2018fault}, spatial distribution in the environment~\cite{culbertson2018decentralized}), significant amount of research has investigated the study of hand-specified controllers designed to solve multi-robot tasks, while overcoming the complexity of coordination~\cite{cortes2017coordinated}. 
Prior work has demonstrated that local interaction rules between the robots can be exploited for the design of task-oriented behaviors, e.g.~\cite{cortes2017coordinated,schwager2011unifying}. However, these task-oriented controllers cannot be directly implemented towards the solution of complex, real-world mission, which usually require an assortment of them (see, for instance~\cite{nagavalli2017automated,pierpaoli2019sequential}, for implementations of this idea). 

\begin{figure}[t]
    \centering
    \includegraphics[width=0.8\columnwidth]{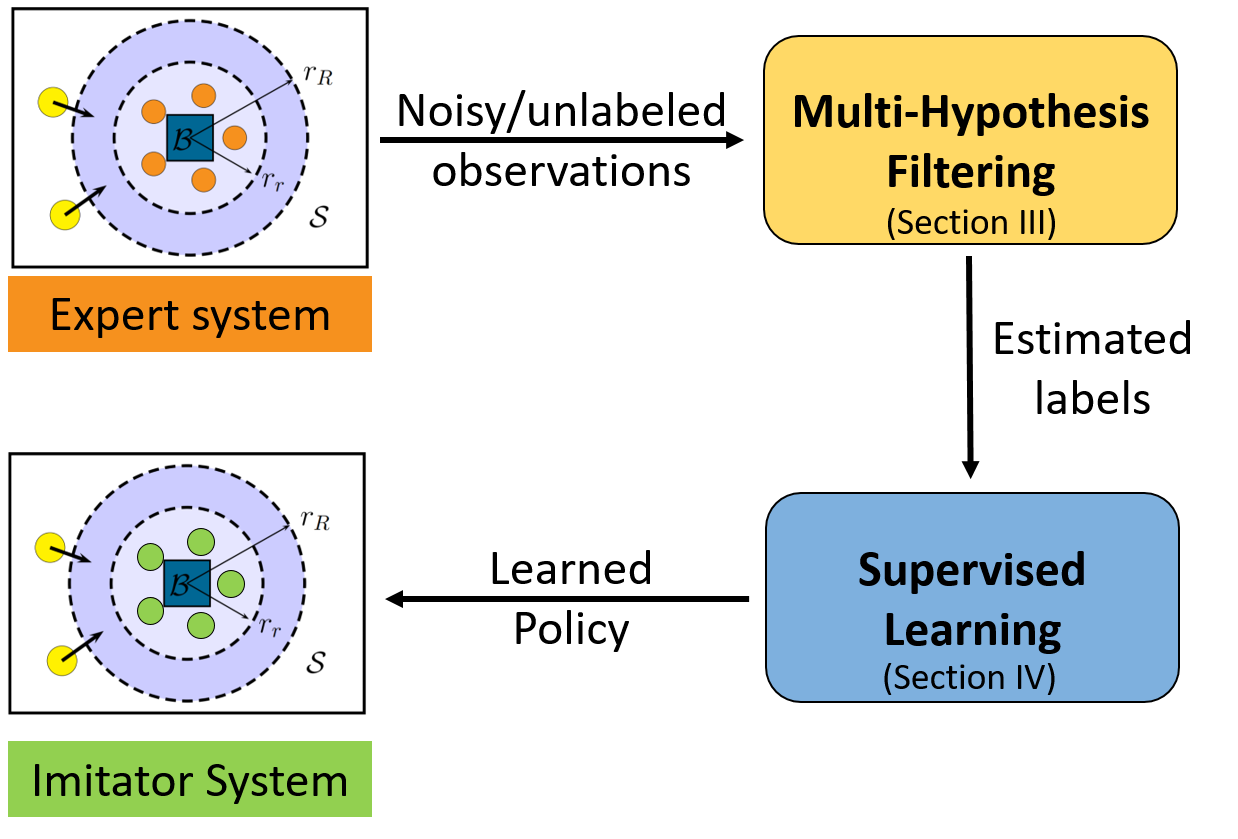}
    \caption{An overview of the proposed approach to learning multi-robot policies from observations.
    }
    \label{fig:overview}
\end{figure}

While it might be possible to enhance the expressiveness of individual controllers through an hand-specified sequential combination of them~\cite{twu2010graph}, a significant amount of human-level intelligence and expertise is required in order to design such controller compositions that are suitable for complex and unknown environments~\cite{nikolaidis2013human}. One way to avoid the burden of requiring significant expertise and design effort is to learn multi-agent policies in an end-to-end manner (e.g., \cite{devin2017learning,xiao2019multi}). However, unlike hand-specified controllers, end-to-end learned policies require vast amounts of data, are not interpretable, or lack low-level control guarantees.

In this work, leveraging the relative merits of both well-understood multi-robot coordinated controllers and learning, we explore a hybrid approach. Specifically, inspired by the paradigm of {\it learning from demonstration}~\cite{argall2009survey,ravichandar2020recent}, we propose a solution to the problem of sequencing multi-robot, task-oriented controllers by leveraging observations of an {\it expert system}. We consider an expert system composed by a team of robots performing a collaborative mission that requires interaction with the environment. Robots belonging to the expert systems are capable of executing a finite set of low-level tasks (e.g., rendezvous, flocking, cyclic pursuit), which are selected by a coordinating, high-level {\it expert policy} based on the state of robots and environment. We consider the expert policy to be a deterministic automated strategy capable of performing the mission with a satisfactory level of performance. Note that our approach is agnostic to whether the expert policy was hand-specified or learned.

We also consider an {\it imitator system}, composed by a team of robots, capable of executing the same list of task-oriented primitive controllers used by the expert system, but unaware of how to compose them in order to solve the mission. The imitator system is capable of collecting noisy observations of both the expert system and the environment as it performances the mission. Note that we do not assume access to demonstrations with annotations of controllers being used. Our objective is a procedure that allows the {\it imitator system} to construct a policy capable of achieving expert-like performances in the same mission.

We propose a two-part framework to learn a controller selection policy from un-annotated demonstrations (see Fig. \ref{fig:overview}). First, based on noisy observations of the expert system's state, we present a state-estimation inference engine that estimates the controller executed by the expert system at any given instant. Second, based on the history of estimated controllers and observed environment variables, we train a neural network to approximate the expert policy. We evaluate the performance of our framework on a perimeter defense task on a team of differential-drive robots. Our results indicate that our approach is capable of successfully learning the expert policy from only noisy and un-annotated demonstrations. 


\subsection{Related Work}
Robotic systems capable of inferring expert behavior have been studied in variety of contexts, such as human-robot interaction (HRI) \cite{wang2013probabilistic,luo2018unsupervised}, autonomous navigation \cite{kelley2008understanding,jain2016recurrent}, and interactive learning \cite{cakmak2012designing, dillmann2004teaching}. For instance, within the context of HRI, inferring human behavior is often referred to as intention inference, wherein the robot attempts to identify the underlying intention or goal of the human partner \cite{ravichandar2016human, ravichandar2018gaze}.  In context of interactive learning, the robot aims to infer what the human partner is teaching \cite{cakmak2012designing}. In these contexts, it is either assumed that labelled training data can be acquired or that it is sufficient to classify the unlabelled data into abstract clusters. In our work, we are interested not only in inferring the expert's behavior, but also learning from such inferences.

Learning from demonstrations (LfD) is a popular paradigm in robotics that provides a plethora of techniques aimed at learning from and imitating expert behavior~\cite{argall2009survey}. A vast majority of techniques, however, rely on labelled demonstrations and solve a supervised learning problem. A few exceptions to this assumption include learning reward functions from unlabelled demonstrations (e.g., \cite{babes2011apprenticeship,liu2018imitation}), and learning manipulation skills from unlabelled videos (e.g., \cite{sermanet2018time}). 

The methods discussed thus far, however, are limited to single-robot tasks and scenarios. In the context of multi-robot systems, prior work has explored identifying joint-intention and shared plans of a group of agents \cite{demiris2007prediction,pei2011parsing}. However, similar to examples in single-robot systems, the inferred information is not utilized to learn control policies. On the other hand, algorithms that learn multi-agent policies (e.g., \cite{natarajan2010multi,vsovsic2017inverse,bhattacharyya2018multi}) rely on labelled expert demonstrations.

As mentioned earlier, a rich body of works exists on task-oriented controllers for multi agent teams. Coordinated controllers based on weighted consensus protocol have been used to achieve, for example, flocking, coverage, formation control, and cyclic pursuit (see~\cite{cortes2017coordinated} and references therein). Examples of alternative methods include the Null Space Methods~\cite{antonelli2008null}, Navigation Functions~\cite{tanner2005towards}, and Model Predictive Control~\cite{richards2004decentralized}. Related to our contribution, solutions to the problem of composing sequences of controllers include formal methods~\cite{kress2018synthesis}, path planning~\cite{nagavalli2017automated}, Finite State Machines~\cite{marino2009behavioral}, Petri Nets~\cite{klavins2000formalism}, Behavior Trees~\cite{colledanchise2014behavior}, and Reinforcement Learning~\cite{mehta2006optimal}. 
We take inspiration from a variety of communities in order to develop a unified framework to learn multi-robot controller selection policies from unlabelled and noisy observations of an expert system.

We note that the idea of learning how to create sequences from a pre-defined list of controllers, rather than learning entirely new controllers, is motivated by the rich literature on coordinated control for multi-agent system~\cite{mesbahi2010graph}. In addition, although beyond the scope of the current work, the use of coordinated controllers makes our framework naturally predisposed for fully distributed implementations. Finally, by focusing on high-level strategies that compose well-understood controllers, we can interpret the actions executed by the robots.


\section{Problem Formulation} \label{sec:problemformulation}

In this section, we introduce the systems and algorithms relevant to our approach and provide a formal definition of our problem.

\textit{\textbf{Expert system}}: Let us consider a team of $N$ robots interacting with an external environment in order to complete of a mission. We denote by $x(k)\in \mathcal{X} \subseteq \mathbb{R}^{nN}$ the state of the robots, where $n$ is number of dimensions necessary to describe the state of each robot. In addition, we assume the external environment to be a deterministic system with unknown dynamics, whose configuration is described by a measurable, discrete-time variable $e(k)\in\mathcal{E} \subseteq \mathbb{R}^m$, for all $k>0$. At each time step, the motion of the robots is governed by a controller, selected from a library of controllers $\mathcal{L} = \{ f_1(x),\dots,f_M(x) \}$, where each element $f_j(\cdot): \mathcal{X} \mapsto \mathcal{X}$ is a continuous function of the state of the robots. At any given time, all robots are driven by the same controller, which excludes the possibility of sub-teams acting in parallel.

\textit{\textbf{Imitator system}}: In addition to the expert system just described, we assume the existence of a second team of $N$ robots that are identical to those in the expert system, whose dynamics are governed by the same library of controllers $\mathcal{L}$. We refer to this second system as {\it imitator system} and we denote its state by $x_I \in \mathcal{X}$. 

\textit{\textbf{Measurement model}}: Both the expert system and imitator system make observations of the environment under the following observation model:
\begin{equation}
    \hat{e}(k) = e(k) + w_e(k)
\end{equation}
where $\hat{e}(k)$ is the state of the environment affected by measurement noise, and $w_e(k)$ is a zero-mean Gaussian white-noise process with known variance. We denote the sequences of environment measurements by $E^k = \{\hat{e}(k),\hat{e}(k-1),\cdots,\hat{e}(0)\}$.

In order to learn from the expert, the imitator also collects noisy observations of the expert's states as follows 
\begin{equation}
    z(k) = h(x(k)) + w_x(k)
\end{equation}
where $h: \mathcal{X} \rightarrow \mathcal{Z}$ is the imitator's measurement function, and $w_x(k)$ is a zero-mean Gaussian white-noise process with known variance. We denote the sequences of observations by $Z^k = \{z(k),z(k-1),\cdots,z(0)\}$.

\textit{\textbf{Expert policy}}:  A supervising deterministic policy $F: \mathcal{X} \times \mathcal{E} \mapsto \mathcal{L}$ selects controllers in $\mathcal{L}$ as function of the states of robots and environment. Then, the discrete-time dynamics of the robots is
\begin{equation}
    x(k+1) = F(x(k),\hat{e}(k)) + v(k),
\end{equation}
where $v(k)$ is a zero-mean Gaussian white-noise processes, with known variance. The policy $F$ selects controllers so that the mission is executed with a measurable mission performance $\Pi^F$. We refer to the system of robots with state $x$ as {\it expert system}, to $F$ as {\it expert's policy}, and to $\Pi^F$ as the {\it expert's mission performance} under the policy $F$. 

\textit{\textbf{Imitator policy}}: We denote the imitator's learned policy as $\Phi: \mathcal{X} \times \mathcal{E} \mapsto \mathcal{L}$, and thus the dynamics of the imitator are
\begin{equation}
    x_I(k+1) = \Phi(x_I(k),\hat{e}(k)) + v(k) \label{eq:imitator_policy}.
\end{equation}

Given access to noisy observations of the expert system and its interaction with the environment, the objective of our work is to learn that imitator policy $\Phi$ such that it imitates the expert policy $F$ and achieves comparable mission performance. 

In the next two sections, we describe the main contribution of this paper, namely a two-stage procedure for the imitator to learn $\Phi$. First, from noisy observation of the expert's state and exact state of the environment, the imitator creates an online estimate of the controllers being executed by the expert. Then, based on these estimates, we propose a policy approximation procedure that allows the imitator to solve the task with performance comparable to the expert's performance on the same task. 

\section{Inferring Controllers from Observations} \label{sec:IMM}
In this section, we describe the Interactive Multiple Model (IMM) technique~\cite{blom1988interacting}, implemented by the imitator to estimate the sequence of controllers $f_j(\cdot)\in\mathcal{L}$ the expert executes, given a history of observations $Z^k$. The choice of this technique is motivated by its popularity and computational efficiency and we insist that the same objective could be achieved with alternative estimation schemes.

\subsection{Interactive Multiple Models Estimator}
The IMM estimator requires a finite list of controllers (or {\it modes}) whose transitions are described by a Markov process. We denote by $p_{ij}$ the process probability of switching from mode $f_j$ to $f_i$, i.e. 
\begin{equation}
    p_{ij}=\text{Prob}(F_k = f_i \,|\, F_{k-1} = f_j).
\end{equation} 
In addition, we define a bank of filters (e.g. Kalman filters, EKF) each corresponding to a controller in the library $\mathcal{L}$. By combining expert's state observations with modes' transition probabilities, the IMM computes 1) the probability of $f_j$ being active at time $k$, which we denote by $\mu_j(k)$ and 2) the estimate of the expert's state $\hat{x}(k|k)$. The estimation technique follows three main steps, which we briefly discuss here for completeness. We refer the reader to~\cite{mazor1998interacting} for a review of IMM implementations and its possible variations. In the following we denote with $[u]_\otimes$ the outer product of vector $u$, defined as $u\,u^T$ and we use the shorthand notation $x(k^-)$ to denote $x(k-1)$.

\paragraph{Interaction}
First, mixing probability are computed by propagating each modes' probability through the Markov process as
\begin{equation*}
  \mu_{i|j}(k^-|k^-) =  \frac{1}{\bar{c}_j}p_{ij}\mu_i(k^-) \quad \forall i,j\in M 
\end{equation*}
where $\bar{c}_j$ is a normalizing factor. Then, we compute the effect of the modes probabilities on previous state estimates and covariance matrix:
\begin{align*}
    \hat{x}_{0j}(k^- | k^-) &= \sum_{i=1}^M \hat{x}_i(k^-|k^-) \, \mu_{i|j}(k^-|k^-)\\
    P_{0j}(k^-|k^-) &= \sum_{i=1}^M \Big[ P_i(k^-|k^-) + \\
    &[\hat{x}_i(k^-|k^-)-\hat{x}_{0j}(k^-|k^-)]_\otimes \Big]\mu_{i|j}(k^-|k^-)
\end{align*}

\paragraph{Filtering}
From noisy measurements $z(k)$ of the expert, the posterior estimates $\hat{x}_j(k|k)$, covariance matrix $P_j(k|k)$, and likelihood $\Lambda_j(k)$ for each mode are computed by applying EKF iterations for each of the known modes. Then, individual mode probabilities are computed as follows:
\begin{equation*}
    \mu_j(k) = \frac{1}{c} \Lambda_j(k) \sum_{i=1}^M p_{ij} \mu_i(k^-)
\end{equation*}
where $c$ is a normalizing factor.

\paragraph{Combination} 
Final state estimates and convariance matrix are then obtained by combining values from each filter weighted by the probability of the corresponding mode
\begin{align*}
\hat{x}(k|k) &= \sum_{j=1}^M \hat{x}_j(k|k) \mu_j(k) \\
P(k|k) &= \sum_{j=1}^M \Big[ P_j(k|k) + [\hat{x}_j(k|k)-\hat{x}(k|k)]_\otimes \Big] \mu_j(k)
\end{align*}

Every time a new observation is collected, the imitator performs an iteration of the procedure just described, from which we obtain the probability distribution for all the controllers in the library. Then, estimated controller being executed by the expert system a time step $k$ is obtained from a maximum a posteriori (MAP) estimate
\begin{equation}
    f^*_k = \argmax_{j \in \{ 1,\dots,M\} }\, \mu_j(k).
\end{equation}

\subsection{Multi-Robot Coordinated Controllers Inference} \label{sec:imm_application}
We demonstrate the implementation of the IMM estimator to the identification of $5$ controllers on a team of $5$ robots. Taking inspiration from the literature on coordinated behaviors for multi-agent systems~\cite{cortes2017coordinated}, we assume each of the controllers in $\mathcal{L}$ to be composed by two terms. The first term corresponds to a coordinated effect between the robots, which is represented by a weighted consensus term~\cite{mesbahi2010graph}. The second term is the agent's individual objective, used to represent, for example, one or more leaders in the team. Each controller can then be described as:
\begin{equation}
    f_j(x) := -L_{w,j}(x) + v_j(x)
\end{equation}
where $L_{w,j}: \mathcal{X} \mapsto \mathcal{X}$ is the weighted Laplacian of the graph $\mathcal{G}_j$ describing the interactions between robots required by controller $j$~\cite{mesbahi2010graph}, while $v_j: \mathcal{X} \mapsto \mathcal{X}$, represents a leader-like controller. We consider $\mathcal{L}_b$ to be composed by the following $5$ behaviors.

\textit{\textbf{{Cyclic pursuit}}}:
\begin{equation*}
u_i = \sum_{j \in \mathcal{N}_i} R(\theta)\,(x_j - x_i) + (x_c-x_i)
\end{equation*}
where $\theta = 2 r\,\sin \frac{\pi}{N}$, $r$ is the radius of the cycle formed by the robots, and $R(\theta)\in SO(2)$. The point $x_c \in \mathcal{X}$ is the center of the cycle and $\mathcal{N}_i$ is the set of neighbors to agent $i$ specified by $\mathcal{G}_j$.

\textit{\textbf{{Leader-follower}}}:
\begin{align*}
	u_i &= \sum_{j \in \mathcal{N}_i} (\|x_i-x_j\|^2-\delta_{f}^2)(x_j - x_i) \quad i=2,\dots,4\\
	u_1 &= \sum_{j \in \mathcal{N}_1} \left( (\|x_1-x_j\|^2-\delta_f^2)(x_j - x_1) \right) + (x_g-x_1)
\end{align*}
where $\delta_f$ is the desired separation between the robots, $x_g \in \mathcal{X}$ represents the leader's goal, and subscript $1$ denotes the leader.

\textit{\textbf{{Circular / Wedge / Star formations}}}:
\begin{align*}
	u_i &= \sum_{j \in \mathcal{N}_i} (\|x_i-x_j\|^2-\bar{\delta}_{ij}^2)(x_j - x_i) \quad i=2,\dots,4\\
	u_1 &= \sum_{j \in \mathcal{N}_i} (\|x_i-x_j\|^2-\bar{\delta}_{ij}^2)(x_j - x_i) + (x_g-x_1)
\end{align*}
where $\bar{\delta}_{ij}\in \mathbb{R}_+$ is the desired separation between robots $i$ and $j$ designed to achieve circle, wedge, and star shapes, while $x_g$ is the formation leader's goal.

\begin{figure}[h]
    \centering
    \includegraphics[width=\columnwidth, trim={0.5cm 0cm 0.5cm 0cm}]{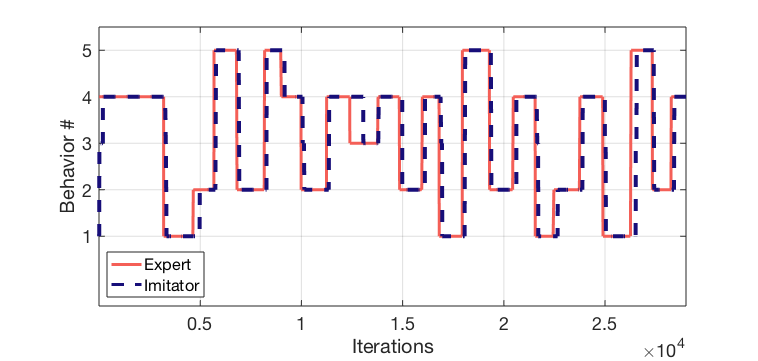}
    \caption{IMM estimation performance for a library of $5$ behaviors. Solid line represents the behavior being executed by the expert system, while the dashed line is the IMM estimate. Controllers order: 1) cyclic pursuit, 2) leader-follower, 3) circle, 4) wedge, and 5) star formations}
    \label{fig:imm_test}
\end{figure}

To illustrate the inference approach, we consider an expert system that randomly selects one of the controllers described above at Poisson instants of time. Results from simulations are reported in Fig.~\ref{fig:imm_test}. As we can observe from results in Fig.~\ref{fig:imm_test}, the IMM correctly estimates the expert's behaviors, provided a minimum sojourn time on each behavior.

\section{Learning to Compose Behaviors} \label{sec:learning}

In this section, we introduce our approach to learn the expert policy $F$. We propose to learn the imitator's policy, denoted by $\Phi$, that approximates the expert policy in mapping the current environmental and estimated robot states onto the appropriate behavior from the controller library $\mathcal{L}$. As noted in (\ref{eq:imitator_policy}), the policy is attempting to capture the general strategy of choosing the behaviors from the expert as opposed to encoding a deterministic sequence of behaviors. In order to train such a policy, we collect the training data from $N$ episodes. Thus, the training data is given by $\mathcal{D} = \{\mathcal{D}^{1}, \mathcal{D}^{2}, \cdots , \mathcal{D}^{N} \}$, where $\mathcal{D}^{i} =  \{{f^*_{k}}^{i}, \hat{e}^{i}(k), \hat{x}^{i}(k)\},\ \forall k$ denotes the data associated with the $i^{\text{th}}$ episode. 

As can be seen from the notation, the training process does not assume access to either the true behavior sequence of the expert systems or the true state of the expert system. That is, the system is trained using inferred quantities $f^{*}$ and $\hat{x}$ provided by the IMM filter. Once the imitator policy $\Phi$ is learned, it can be utilized to compose the individual behaviors of the imitator system in order to solve the mission. 

To illustrate our approach, we utilize a neural network (NN) to represent the imitator's policy. The parameters of the network (i.e. weights) are trained using the standard back propagation algorithm. Note that the choice of neural networks over other models is motivated by its universal approximation capability~\cite{cybenko1989approximation} and is not central to our framework. Any function approximation method capable of sufficiently capturing the expert policy would be applicable.

\section{Experimental Validation} \label{sec:experiment}
In this section, we implement the policy approximation technique described in the previous sections, on the observations of an expert system performing a {\it perimeter protection} mission. During expert's execution of the mission, the imitator collects observations of the expert's state and environment's state, in order to estimate what controllers are used by the experts for different configurations of the environment. Then, using data from the estimation process, we train the imitator's policy, which is finally executed by the imitator on the same scenario.

\subsection{Perimeter Defense Scenario} \label{sec:policies}
Perimeter protection can be considered both as a stand-alone scenario or as an abstraction of sub-tasks common to many multi-robot missions. As such, it is an ideal testing scenarios for multi-agent control protocols~\cite{shishika2018local}. In this scenario, the expert team is tasked with defending a region while an adversarial team, representing the environment, tries to intrude. The intruders' objective is to reach the innermost region (dark blue), which we denote by $\mathcal{B}$. The objective of the defenders is to prevent intruders from reaching the protected area $\mathcal{B}$. To this end, the policy governing the defending agents reactively chooses the appropriate controller in $\mathcal{L}$.


\begin{figure*}[t]
	\begin{center}
		\includegraphics[trim={2cm 0.5cm 2cm 0cm}, width=0.65\columnwidth]{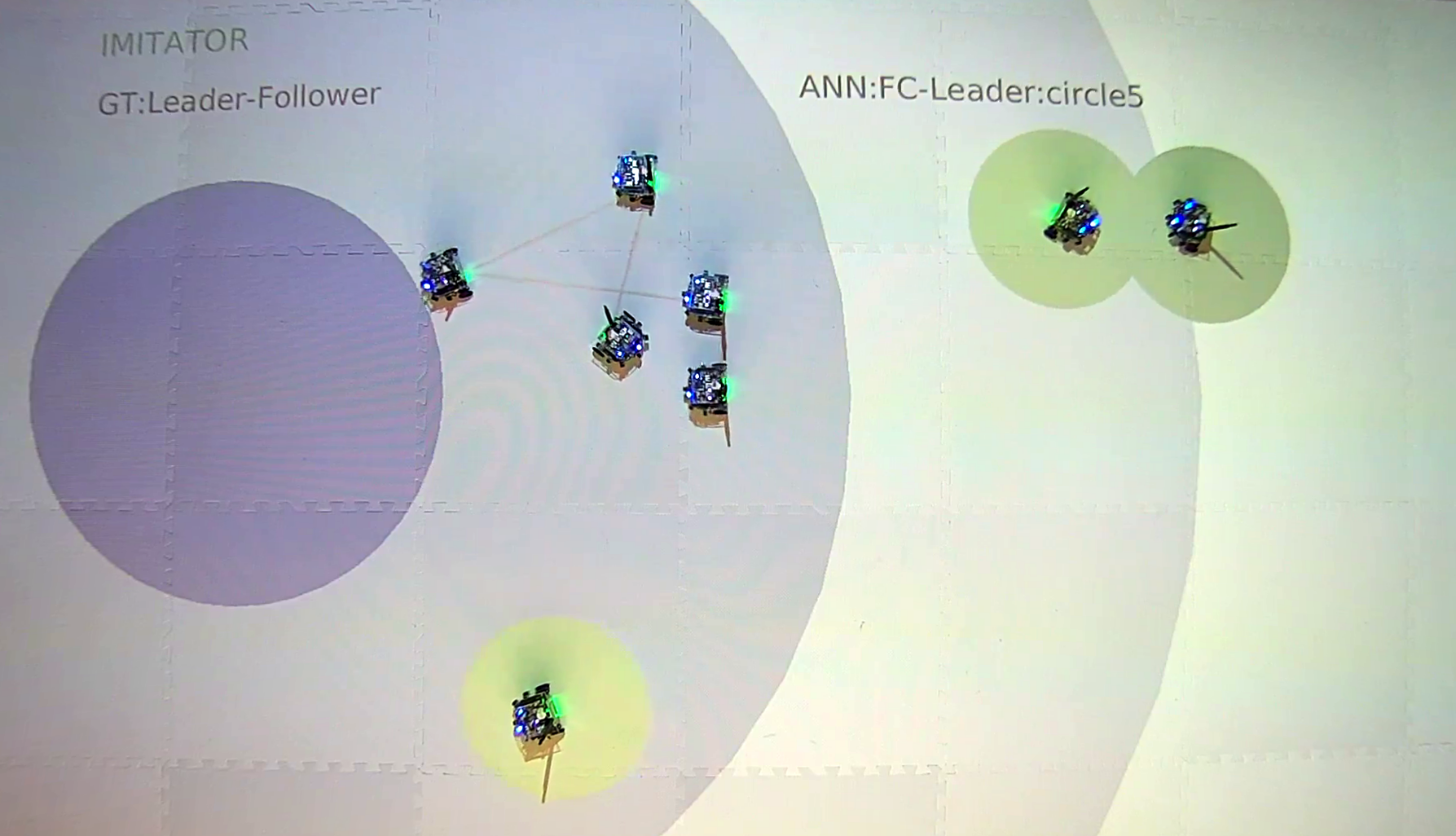}~
		\includegraphics[trim={2cm 0.5cm 2cm 0cm},width=0.65\columnwidth]{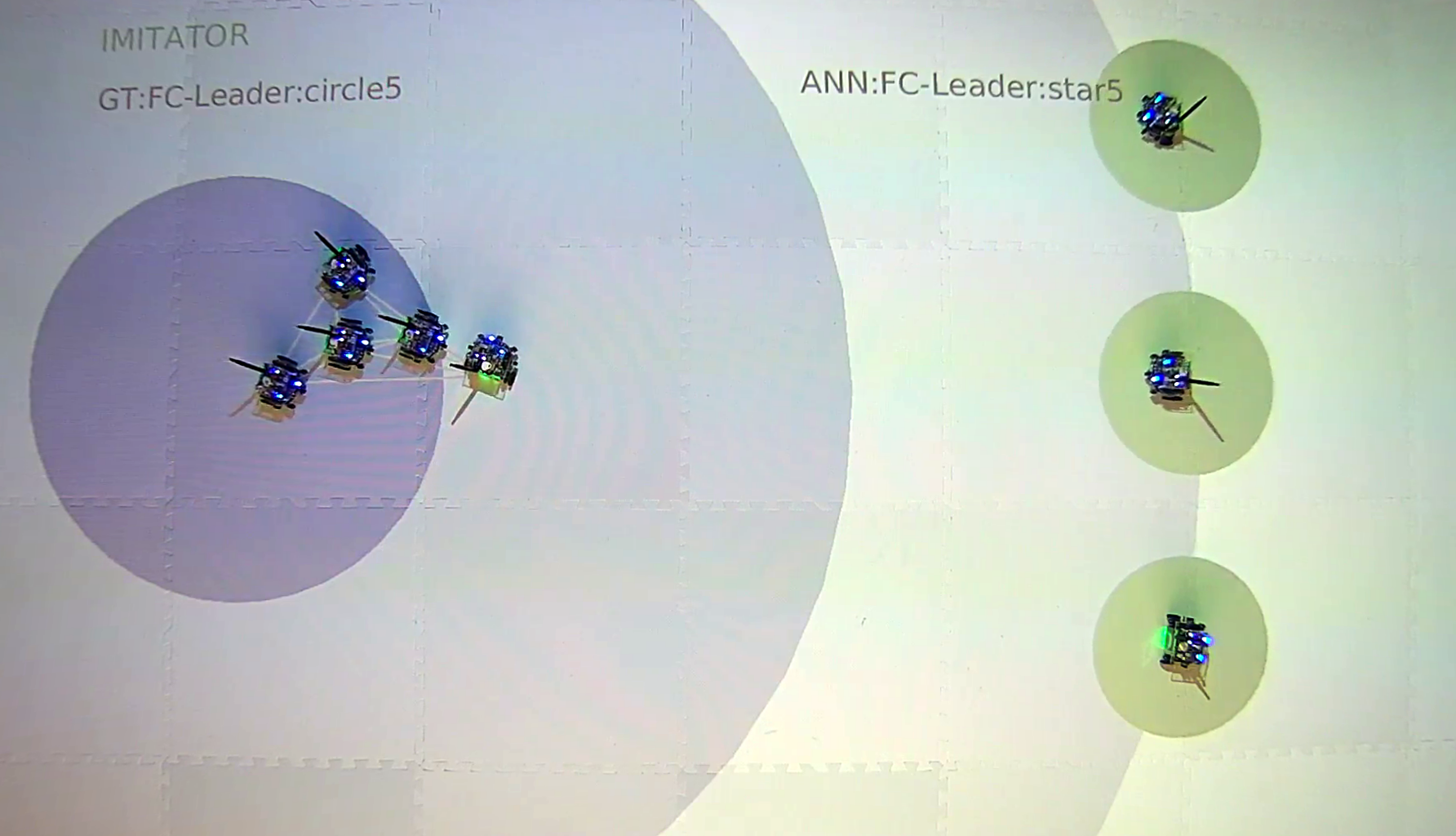}~
		\includegraphics[trim={2cm 0.5cm 2cm 0cm},width=0.65\columnwidth]{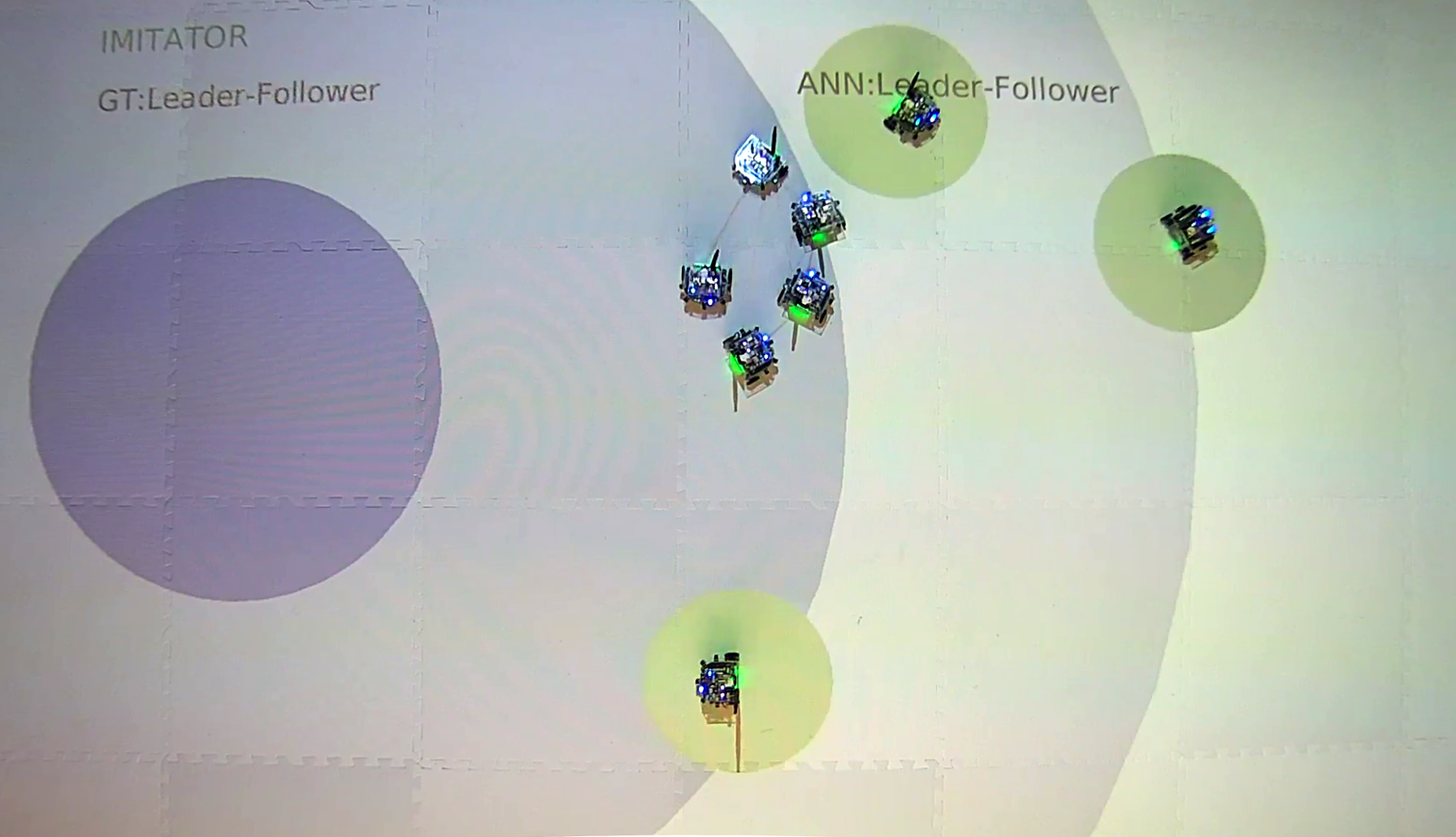}
		\caption{Perimeter protection scenario executed on the Robotarium. Intruders are distinguishable by the yellow disc representing their capture area. The three circular regions in different shades of blue, from dark to light, represent $\mathcal{B}$, $r_r$, and $r_R$ respectively. Text in figure denotes the controller from the expert policy (left) and the controller from the learned policy (right).\label{fig:box_simu} 
		}
	\end{center}
\end{figure*}

\subsubsection{Intruders' strategy} 
\begin{figure}[H]
    \centering
    \includegraphics[width=\columnwidth]{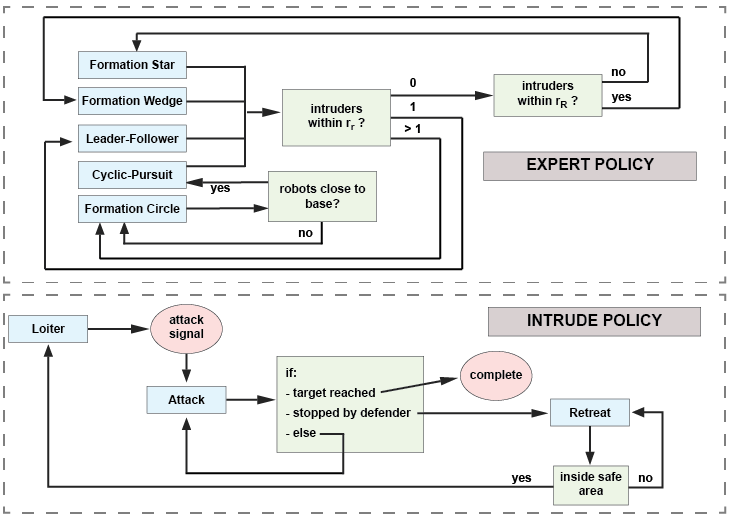}
    \caption{Expert's (top) and intruders' (bottom) policies.
    }
    \label{fig:policies}
\end{figure}

We assume three intruders, with collective state $e(k)=[e_1(k),e_2(k),e_3(k)]^T$, each described at any given time by three possible modes, namely {\sc loiter}, {\sc attack}, {\sc retreat}. In addition, each intruder is equipped with a circular protected area of radius $\varepsilon_{d}$ around it. The intruder's strategy is illustrated in Fig. \ref{fig:policies}. At initial time, intruders start at random points inside region $\mathcal{S}$, in {\sc loiter} mode. At Poisson distributed instants of time, a uniformly distributed random number of intruders are selected and switched to {\sc attack} mode. Intruders in {\sc attack} mode proceed towards the center of region $\mathcal{B}$. During attacks, if any of the intruders encounter a defending robot within distance less than $\varepsilon_d$, the intruder switches its state to {\sc retreat} and moves back to uniformly selected points in region $\mathcal{S}$. 

\subsubsection{Expert Policy} 
The policy employed by the defending robots acts by selecting controllers in the library $\mathcal{L}$ (see Fig. \ref{fig:policies}). Here, we assume $\mathcal{L}$ to be composed of the same controllers described in Section~\ref{sec:imm_application}, namely {\sc leader-follower}, {\sc cyclic-pursuit}, and three different formation assembling protocols {\sc wedge, circular}, and {\sc star}. When none of the intruders are within distance $r_R$, defending robots execute the {\sc circle}  formation centered inside region $\mathcal{B}$. If one intruder is within distance $r_r$, defenders switch to the {\sc leader-follower} controller. If two or more intruders are within distance $r_r$, defenders first move closer to $\mathcal{B}$ following the {\sc circle formation} controller then switch to {\sc cyclic pursuit}. Then, whether one or more intruders are between distances $r_R$ and $r_r$ robots switch to either {\sc star} or {\sc wedge} formations respectively.

\subsection{Experimental Design}
In this section, we describe the details and design choices associated with our experiments.

\subsubsection{Simulation and Experimentation Platform}
We realized both a simulated and a physical implementation of the perimeter protection scenario on the Robotarium~\cite{wilson2020robotarium}, which includes a number of features of the real robots, e.g., differential drive kinematics and collision avoidance through barrier control functions. In our experiments, we define an episode as the time between initiations of two consecutive intruders' attacks. In total, the data collected in simulation produced the demonstration set $\mathcal{D}$ consisting of $1000$ episodes. 

\subsubsection{Imitator variations} 
In order to evaluate the performance of the proposed approach, we learn the imitator policy under two distinct conditions:
\begin{flushitemize}
    \item $\Phi^{GT}: \mathcal{X} \times \mathcal{E} \mapsto \mathcal{L}$; Policy learned from the controller sequence generated by the expert policy, ground-truth robot state, and noisy environmental state ($\mathcal{D}^{i}_{GT} = \{{f_{k}}^{i}, \hat{e}^{i}(k), x^{i}(k)\},\ \forall i,k$)
    \item $\Phi^{IMM}: \mathcal{X} \times \mathcal{E} \mapsto \mathcal{L}$; Policy learned from the sequence of estimated controllers, estimated robot states, and noisy environmental states ($\mathcal{D}^{i}_{IMM} = \{{f^*_{k}}^{i}, \hat{e}^{i}(k), \hat{x}^{i}(k)\},\ \forall i,k$) 
\end{flushitemize}
Note that the imitator variation $\Phi^{GT}$ represents the ``best-case" imitation learning scenario in which ground-truth labeled demonstrations are available.

\subsubsection{Training details}
The two policies described above are represented using a multi-layer neural network with two 32 neuron, fully connected layers with hyperbolic tangent activation and an output classification layer with softmax activation to classify between the $5$ control modes. For each policy, the corresponding demonstration data were split into $80\%$ training ($800$ episodes) and $20\%$ validation data ($200$ episodes). 

\subsubsection{Metrics}
We measure the performance of each imitator variation described above in terms of the following metrics:

{\it \textbf{Inference accuracy}}: To evaluate the performance of the inference algorithm, we report the fraction of instants in which our approach successfully inferred the correct expert behavior.

{\it \textbf{Learning accuracy}}: We report training and validation accuracy of the neural networks associated with both $\Phi^{GT}$ and $\Phi^{IMM}$. The accuracy for $\Phi^{GT}$ ($\Phi^{IMM}$) is computed by comparing the predicted behaviors and the behaviors found in the validation set of $\mathcal{D}^{i}_{GT}$ ($\mathcal{D}^{i}_{IMM}$).

{\it \textbf{Imitation performance}}: To investigate how well the imitator has learned the expert system's policy from noisy data, we compute the imitation accuracy of the learned policy by comparing their predicted behavior to that of the expert policy.

{\it \textbf{Mission performance}}: To evaluate the performance on the actual mission, we compute the ratio of the number of attacks that were successfully thwarted by the defending robots to the total number of initiated attacks. 

\subsection{Results and Discussion}

\subsubsection{How well can we infer expert behavior?} 

The performance of the IMM in estimating controllers executed by the expert during the perimeter defense scenario is reported in Fig.~\ref{fig:IMM_perf_peri}. We note that, in this scenario, accurately estimating {\sc circle-formation}, {\sc wedge-formation},  and {\sc leader-follower} is particularly challenging since, by definition of the expert policy, these controllers are executed for a limited period of time, and as a result, the estimator is unlikely to acquire sufficient evidence before the change in underlying behavior. Nevertheless, the estimator is able to achieve an overall accuracy of $75\%$.

\subsubsection{How well do the NNs approximate the available data?}
Training results for the two different policies are reported in Fig.~\ref{fig:trainingPerformance}. As one would expect, training with ground truth data ($\Phi^{GT}$) results in high learning accuracy in the validation set. In support of our framework, the IMM-based policy ($\Phi^{IMM}$) is also shown to exhibit similarly high learning accuracy. Thus, in both cases, the corresponding neural network is capable of accurately learning the available data.

\subsubsection{How well can we learn the expert policy from noisy data?}
\begin{figure}[h]
    \centering
    \includegraphics[width=\columnwidth]{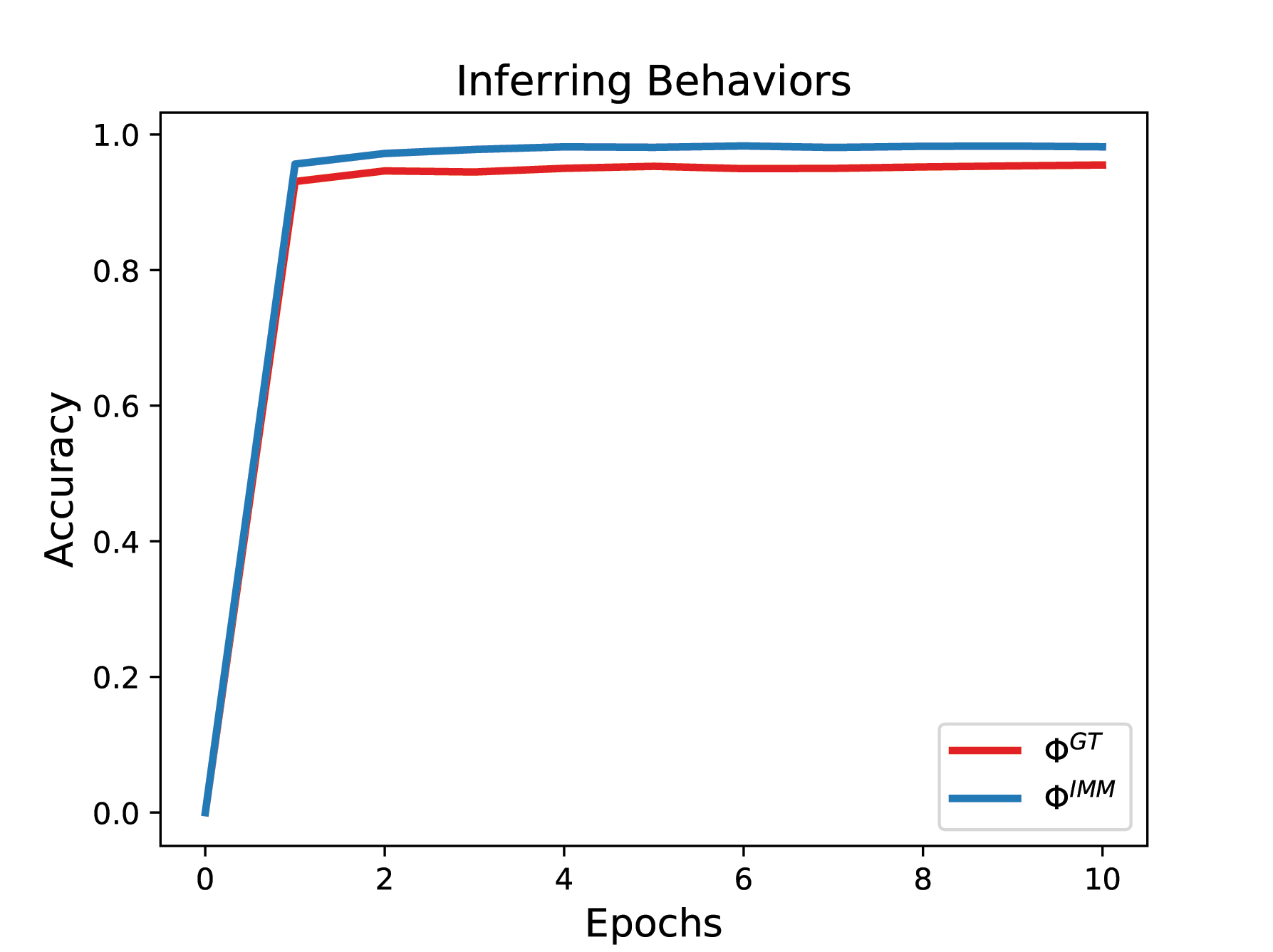}
    \caption{Learning accuracy on the validation set  for the IMM and GT imitator variations.
    }
    \label{fig:trainingPerformance}
\end{figure}

\begin{figure}[h]
    \centering
    \includegraphics[width=\columnwidth, trim = {1cm 0cm 0cm 0cm}]{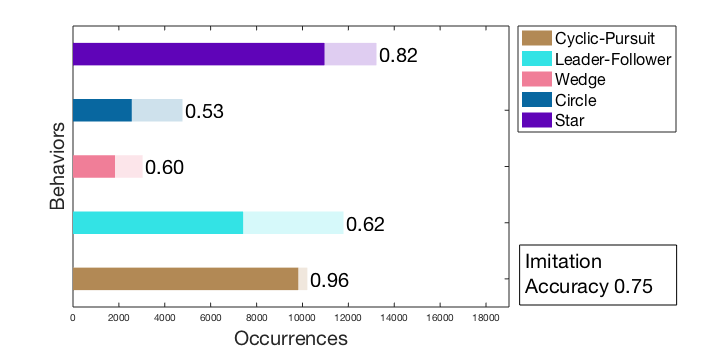}
    \caption{Imitation performance in the perimeter protection scenario. Translucent bars represent the total number of behavior occurrences in the ground truth data while each solid bar represents the number of times $\Phi^{IMM}$ predicted the correct behavior. \label{fig:ImitPerf}}
\end{figure}

In addition to training and validating of the learned policies, we evaluate how effectively the imitators learn to approximate the expert's policies ($F$). In particular, as shown in Fig.~\ref{fig:ImitPerf}, over a sequence of $120$ episodes, the behavior chosen by the policy $\Phi^{IMM}$ matched what would have been chosen by the expert policy $75\%$ of the time on average. Indeed, the imitation performance is affected by both the inference accuracy and learning accuracy. We note that the imitation performance mirrors that of inference. This is likely due to the high learning performance as depicted in Fig. \ref{fig:trainingPerformance}. 


\begin{figure}[h]
    \centering
    \includegraphics[width=\columnwidth, trim = {1cm 0cm 0cm 0cm}]{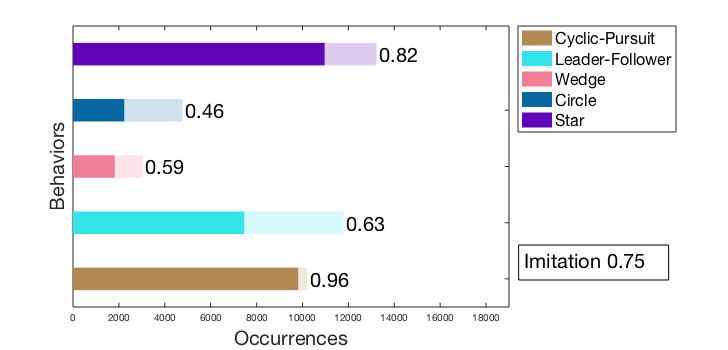}
    \caption{Inference accuracy in the perimeter protection scenario. Translucent bars represent the total number of behavior occurrences while each solid bar represents the number of times the corresponding behavior was correctly estimated.\label{fig:IMM_perf_peri}}
\end{figure}

\subsubsection{What is the effect of the learned policies on mission performance?}

\begin{figure}[ht]
    \centering
    \includegraphics[width=\columnwidth]{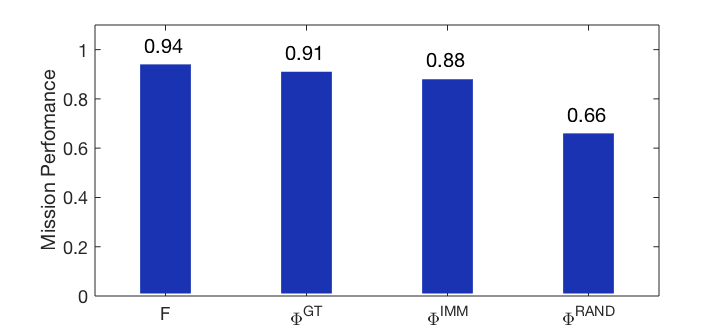}
    \caption{
    Mission performance of the expert policy ($F$), imitator policies ($\Phi^{GT}$ and $\Phi^{IMM}$), and the random policy ($\Phi^{RAND}$).
    \label{fig:intruder_perf} 
    }
\end{figure}

Finally, the mission performance of the learned policies was evaluated in the perimeter defense scenario on a sequence of $120$ episodes.
As shown in Fig.~\ref{fig:intruder_perf}, we compare the performance $\Pi$ of the expert policy $F$, the imitator's policies $\Phi^{GT}$ and $\Phi^{IMM}$, and a random policy. For the random policy, behaviors are switched uniformly at random, at exponentially distributed instants of times\footnote{The switching intervals obtained by fitting an exponential probability density function to the switching intervals observed from the expert policy}.
Despite the limited performance in estimating three of the behaviors as shown in Fig.~\ref{fig:IMM_perf_peri}, we observe that the imitator's mission performance does not deviate significantly from that of the expert. This is explained by the limited impact of the misclassified behaviors on the mission performance.


\section{Conclusions}
In this paper we investigated the problem of learning a multi-agent behaviors selection policy from un-annotated observations of an expert executing a mission. Continuous observations of the expert’s positions are first converted by the imitator into a list of estimated controllers. Then, the history of estimated controllers, robots' and environment's configurations are used to learn the approximated policy. The work presented in this paper is a step towards the development of observation-based learning frameworks for multi-robot systems. In particular, our learning-from-observations framework could be extended to the solution of other forms of multi-robot selection processes, such as task-assignments or heterogeneous robots team-composition, which are notoriously computationally hard to solve with traditional techniques.

\bibliographystyle{IEEEtran}
\bibliography{IEEEabrv,main.bib}

\begin{thebibliography}{10}
\providecommand{\url}[1]{#1}
\csname url@samestyle\endcsname
\providecommand{\newblock}{\relax}
\providecommand{\bibinfo}[2]{#2}
\providecommand{\BIBentrySTDinterwordspacing}{\spaceskip=0pt\relax}
\providecommand{\BIBentryALTinterwordstretchfactor}{4}
\providecommand{\BIBentryALTinterwordspacing}{\spaceskip=\fontdimen2\font plus
\BIBentryALTinterwordstretchfactor\fontdimen3\font minus
  \fontdimen4\font\relax}
\providecommand{\BIBforeignlanguage}[2]{{%
\expandafter\ifx\csname l@#1\endcsname\relax
\typeout{** WARNING: IEEEtran.bst: No hyphenation pattern has been}%
\typeout{** loaded for the language `#1'. Using the pattern for}%
\typeout{** the default language instead.}%
\else
\language=\csname l@#1\endcsname
\fi
#2}}
\providecommand{\BIBdecl}{\relax}
\BIBdecl

\bibitem{ramachandran2019resilience}
R.~K. Ramachandran, J.~A. Preiss, and G.~S. Sukhatme, ``Resilience by
  reconfiguration: Exploiting heterogeneity in robot teams,'' \emph{arXiv
  preprint arXiv:1903.04856}, 2019.

\bibitem{pierpaoli2018fault}
P.~Pierpaoli, D.~Sauter, and M.~Egerstedt, ``Fault tolerant control for
  networked mobile robots,'' in \emph{2018 IEEE Conference on Control
  Technology and Applications (CCTA)}.\hskip 1em plus 0.5em minus 0.4em\relax
  IEEE, 2018, pp. 374--379.

\bibitem{culbertson2018decentralized}
P.~Culbertson and M.~Schwager, ``Decentralized adaptive control for
  collaborative manipulation,'' in \emph{2018 IEEE International Conference on
  Robotics and Automation (ICRA)}.\hskip 1em plus 0.5em minus 0.4em\relax IEEE,
  2018, pp. 278--285.

\bibitem{cortes2017coordinated}
J.~Cort{\'e}s and M.~Egerstedt, ``Coordinated control of multi-robot systems: A
  survey,'' \emph{SICE Journal of Control, Measurement, and System
  Integration}, vol.~10, no.~6, pp. 495--503, 2017.

\bibitem{schwager2011unifying}
M.~Schwager, D.~Rus, and J.-J. Slotine, ``Unifying geometric, probabilistic,
  and potential field approaches to multi-robot deployment,'' \emph{The
  International Journal of Robotics Research}, vol.~30, no.~3, pp. 371--383,
  2011.

\bibitem{nagavalli2017automated}
S.~Nagavalli, N.~Chakraborty, and K.~Sycara, ``Automated sequencing of swarm
  behaviors for supervisory control of robotic swarms,'' in \emph{2017 IEEE
  International Conference on Robotics and Automation (ICRA)}.\hskip 1em plus
  0.5em minus 0.4em\relax IEEE, 2017, pp. 2674--2681.

\bibitem{pierpaoli2019sequential}
P.~Pierpaoli, A.~Li, M.~Srinivasan, X.~Cai, S.~Coogan, and M.~Egerstedt, ``A
  sequential composition framework for coordinating multi-robot behaviors,''
  \emph{arXiv preprint arXiv:1907.07718}, 2019.

\bibitem{twu2010graph}
P.~Twu, P.~Martin, and M.~Egerstedt, ``Graph process specifications for hybrid
  networked systems,'' \emph{IFAC Proceedings Volumes}, vol.~43, no.~12, pp.
  65--70, 2010.

\bibitem{nikolaidis2013human}
S.~Nikolaidis and J.~Shah, ``Human-robot cross-training: computational
  formulation, modeling and evaluation of a human team training strategy,'' in
  \emph{Proceedings of the 8th ACM/IEEE international conference on Human-robot
  interaction}.\hskip 1em plus 0.5em minus 0.4em\relax IEEE Press, 2013, pp.
  33--40.

\bibitem{devin2017learning}
C.~Devin, A.~Gupta, T.~Darrell, P.~Abbeel, and S.~Levine, ``Learning modular
  neural network policies for multi-task and multi-robot transfer,'' in
  \emph{2017 IEEE International Conference on Robotics and Automation
  (ICRA)}.\hskip 1em plus 0.5em minus 0.4em\relax IEEE, 2017, pp. 2169--2176.

\bibitem{xiao2019multi}
Y.~Xiao, J.~Hoffman, T.~Xia, and C.~Amato, ``Multi-robot deep reinforcement
  learning with macro-actions,'' \emph{arXiv preprint arXiv:1909.08776}, 2019.

\bibitem{argall2009survey}
B.~D. Argall, S.~Chernova, M.~Veloso, and B.~Browning, ``A survey of robot
  learning from demonstration,'' \emph{Robotics and autonomous systems},
  vol.~57, no.~5, pp. 469--483, 2009.

\bibitem{ravichandar2020recent}
H.~Ravichandar, A.~S. Polydoros, S.~Chernova, and A.~Billard, ``Recent advances
  in robot learning from demonstration,'' \emph{Annual Review of Control,
  Robotics, and Autonomous Systems}, vol.~3, 2020.

\bibitem{wang2013probabilistic}
Z.~Wang, K.~M{\"u}lling, M.~P. Deisenroth, H.~Ben~Amor, D.~Vogt,
  B.~Sch{\"o}lkopf, and J.~Peters, ``Probabilistic movement modeling for
  intention inference in human--robot interaction,'' \emph{The International
  Journal of Robotics Research}, vol.~32, no.~7, pp. 841--858, 2013.

\bibitem{luo2018unsupervised}
R.~Luo, R.~Hayne, and D.~Berenson, ``Unsupervised early prediction of human
  reaching for human--robot collaboration in shared workspaces,''
  \emph{Autonomous Robots}, vol.~42, no.~3, pp. 631--648, 2018.

\bibitem{kelley2008understanding}
R.~Kelley, A.~Tavakkoli, C.~King, M.~Nicolescu, M.~Nicolescu, and G.~Bebis,
  ``Understanding human intentions via hidden markov models in autonomous
  mobile robots,'' in \emph{Proceedings of the 3rd ACM/IEEE international
  conference on Human robot interaction}.\hskip 1em plus 0.5em minus
  0.4em\relax ACM, 2008, pp. 367--374.

\bibitem{jain2016recurrent}
A.~Jain, A.~Singh, H.~S. Koppula, S.~Soh, and A.~Saxena, ``Recurrent neural
  networks for driver activity anticipation via sensory-fusion architecture,''
  in \emph{2016 IEEE International Conference on Robotics and Automation
  (ICRA)}.\hskip 1em plus 0.5em minus 0.4em\relax IEEE, 2016, pp. 3118--3125.

\bibitem{cakmak2012designing}
M.~Cakmak and A.~L. Thomaz, ``Designing robot learners that ask good
  questions,'' in \emph{Proceedings of the seventh annual ACM/IEEE
  international conference on Human-Robot Interaction}.\hskip 1em plus 0.5em
  minus 0.4em\relax ACM, 2012, pp. 17--24.

\bibitem{dillmann2004teaching}
R.~Dillmann, ``Teaching and learning of robot tasks via observation of human
  performance,'' \emph{Robotics and Autonomous Systems}, vol.~47, no. 2-3, pp.
  109--116, 2004.

\bibitem{ravichandar2016human}
H.~C. Ravichandar and A.~P. Dani, ``Human intention inference using
  expectation-maximization algorithm with online model learning,'' \emph{IEEE
  Transactions on Automation Science and Engineering}, vol.~14, no.~2, pp.
  855--868, 2016.

\bibitem{ravichandar2018gaze}
H.~C. Ravichandar, A.~Kumar, and A.~Dani, ``Gaze and motion information fusion
  for human intention inference,'' \emph{International Journal of Intelligent
  Robotics and Applications}, vol.~2, no.~2, pp. 136--148, 2018.

\bibitem{babes2011apprenticeship}
M.~Babes, V.~Marivate, K.~Subramanian, and M.~L. Littman, ``Apprenticeship
  learning about multiple intentions,'' in \emph{Proceedings of the 28th
  International Conference on Machine Learning (ICML-11)}, 2011, pp. 897--904.

\bibitem{liu2018imitation}
Y.~Liu, A.~Gupta, P.~Abbeel, and S.~Levine, ``Imitation from observation:
  Learning to imitate behaviors from raw video via context translation,'' in
  \emph{2018 IEEE International Conference on Robotics and Automation
  (ICRA)}.\hskip 1em plus 0.5em minus 0.4em\relax IEEE, 2018, pp. 1118--1125.

\bibitem{sermanet2018time}
P.~Sermanet, C.~Lynch, Y.~Chebotar, J.~Hsu, E.~Jang, S.~Schaal, S.~Levine, and
  G.~Brain, ``Time-contrastive networks: Self-supervised learning from video,''
  in \emph{2018 IEEE International Conference on Robotics and Automation
  (ICRA)}.\hskip 1em plus 0.5em minus 0.4em\relax IEEE, 2018, pp. 1134--1141.

\bibitem{demiris2007prediction}
Y.~Demiris, ``Prediction of intent in robotics and multi-agent systems,''
  \emph{Cognitive processing}, vol.~8, no.~3, pp. 151--158, 2007.

\bibitem{pei2011parsing}
M.~Pei, Y.~Jia, and S.-C. Zhu, ``Parsing video events with goal inference and
  intent prediction,'' in \emph{2011 International Conference on Computer
  Vision}.\hskip 1em plus 0.5em minus 0.4em\relax IEEE, 2011, pp. 487--494.

\bibitem{natarajan2010multi}
S.~Natarajan, G.~Kunapuli, K.~Judah, P.~Tadepalli, K.~Kersting, and J.~Shavlik,
  ``Multi-agent inverse reinforcement learning,'' in \emph{2010 Ninth
  International Conference on Machine Learning and Applications}.\hskip 1em
  plus 0.5em minus 0.4em\relax IEEE, 2010, pp. 395--400.

\bibitem{vsovsic2017inverse}
A.~{\v{S}}o{\v{s}}i{\'c}, W.~R. KhudaBukhsh, A.~M. Zoubir, and H.~Koeppl,
  ``Inverse reinforcement learning in swarm systems,'' in \emph{Proceedings of
  the 16th Conference on Autonomous Agents and MultiAgent Systems}.\hskip 1em
  plus 0.5em minus 0.4em\relax International Foundation for Autonomous Agents
  and Multiagent Systems, 2017, pp. 1413--1421.

\bibitem{bhattacharyya2018multi}
R.~P. Bhattacharyya, D.~J. Phillips, B.~Wulfe, J.~Morton, A.~Kuefler, and M.~J.
  Kochenderfer, ``Multi-agent imitation learning for driving simulation,'' in
  \emph{2018 IEEE/RSJ International Conference on Intelligent Robots and
  Systems (IROS)}.\hskip 1em plus 0.5em minus 0.4em\relax IEEE, 2018, pp.
  1534--1539.

\bibitem{antonelli2008null}
G.~Antonelli, F.~Arrichiello, and S.~Chiaverini, ``The null-space-based
  behavioral control for autonomous robotic systems,'' \emph{Intelligent
  Service Robotics}, vol.~1, no.~1, pp. 27--39, 2008.

\bibitem{tanner2005towards}
H.~G. Tanner and A.~Kumar, ``Towards decentralization of multi-robot navigation
  functions,'' in \emph{Proceedings of the 2005 IEEE International Conference
  on Robotics and Automation}.\hskip 1em plus 0.5em minus 0.4em\relax IEEE,
  2005, pp. 4132--4137.

\bibitem{richards2004decentralized}
A.~Richards and J.~How, ``Decentralized model predictive control of cooperating
  \sc{UAV}s,'' in \emph{2004 43rd IEEE Conference on Decision and Control
  (CDC)(IEEE Cat. No. 04CH37601)}, vol.~4.\hskip 1em plus 0.5em minus
  0.4em\relax IEEE, 2004, pp. 4286--4291.

\bibitem{kress2018synthesis}
H.~Kress-Gazit, M.~Lahijanian, and V.~Raman, ``Synthesis for robots: Guarantees
  and feedback for robot behavior,'' \emph{Annual Review of Control, Robotics,
  and Autonomous Systems}, vol.~1, pp. 211--236, 2018.

\bibitem{marino2009behavioral}
A.~Marino, L.~Parker, G.~Antonelli, and F.~Caccavale, ``Behavioral control for
  multi-robot perimeter patrol: A finite state automata approach,'' in
  \emph{2009 IEEE International Conference on Robotics and Automation}.\hskip
  1em plus 0.5em minus 0.4em\relax IEEE, 2009, pp. 831--836.

\bibitem{klavins2000formalism}
E.~Klavins and D.~E. Koditschek, ``A formalism for the composition of
  concurrent robot behaviors,'' in \emph{Proceedings 2000 ICRA. Millennium
  Conference. IEEE International Conference on Robotics and Automation.
  Symposia Proceedings (Cat. No. 00CH37065)}, vol.~4.\hskip 1em plus 0.5em
  minus 0.4em\relax IEEE, 2000, pp. 3395--3402.

\bibitem{colledanchise2014behavior}
M.~Colledanchise and P.~{\"O}gren, ``How behavior trees modularize robustness
  and safety in hybrid systems,'' in \emph{2014 IEEE/RSJ International
  Conference on Intelligent Robots and Systems}.\hskip 1em plus 0.5em minus
  0.4em\relax IEEE, 2014, pp. 1482--1488.

\bibitem{mehta2006optimal}
T.~R. Mehta and M.~Egerstedt, ``An optimal control approach to mode generation
  in hybrid systems,'' \emph{Nonlinear Analysis: Theory, Methods \&
  Applications}, vol.~65, no.~5, pp. 963--983, 2006.

\bibitem{mesbahi2010graph}
M.~Mesbahi and M.~Egerstedt, \emph{Graph theoretic methods in multiagent
  networks}.\hskip 1em plus 0.5em minus 0.4em\relax Princeton University Press,
  2010, vol.~33.

\bibitem{blom1988interacting}
H.~A. Blom and Y.~Bar-Shalom, ``The interacting multiple model algorithm for
  systems with markovian switching coefficients,'' \emph{IEEE transactions on
  Automatic Control}, vol.~33, no.~8, pp. 780--783, 1988.

\bibitem{mazor1998interacting}
E.~Mazor, A.~Averbuch, Y.~Bar-Shalom, and J.~Dayan, ``Interacting multiple
  model methods in target tracking: a survey,'' \emph{IEEE Transactions on
  aerospace and electronic systems}, vol.~34, no.~1, pp. 103--123, 1998.

\bibitem{cybenko1989approximation}
G.~Cybenko, ``Approximation by superpositions of a sigmoidal function,''
  \emph{Mathematics of control, signals and systems}, vol.~2, no.~4, pp.
  303--314, 1989.

\bibitem{shishika2018local}
D.~Shishika and V.~Kumar, ``Local-game decomposition for multiplayer
  perimeter-defense problem,'' in \emph{2018 IEEE Conference on Decision and
  Control (CDC)}.\hskip 1em plus 0.5em minus 0.4em\relax IEEE, 2018, pp.
  2093--2100.

\bibitem{wilson2020robotarium}
S.~Wilson, P.~Glotfelter, L.~Wang, S.~Mayya, G.~Notomista, M.~Mote, and
  M.~Egerstedt, ``The robotarium: Globally impactful opportunities, challenges,
  and lessons learned in remote-access, distributed control of multirobot
  systems,'' \emph{IEEE Control Systems Magazine}, vol.~40, no.~1, pp. 26--44,
  2020.

\end{thebibliography}

\end{document}